# अंग्रेजी - हिन्दी मशीनी अनुवाद का मानव और स्वचालित मूल्यांकन

निशीथ जोशी[*], हेमंत दरबारी[**], इति माथुर[*]

[*] आपाजी संस्थान, वनस्थली विद्यापीठ, राजस्थान

[**] प्रगत संगणन विकास केंद्र, पुणे, महाराष्ट्र

*सारांश:* मशीनी अनुवाद में अनुसन्धान करीब साठ साल से चल रहा है । इस विषय के विकास के लिए, दुनिया भर में वैज्ञानिक नितनयी तकनीक का विकास कर रहे है । परिणामस्वरूप हमें कई नए स्वचालित मशीनी अनुवादक भी मिले है । एक मशीनी अनुवादक के प्रबंधक के लिए यह पता करना बहुत महत्वपूर्ण है की उसकी प्रणाली में संशोधनों के बाद कितना सुधार हुआ । इसी कारणवश मशीनी अनुवादक के मूल्यांकन की आवश्कता पड़ी । इस लेख में हम कुछ मशीनी अनुवादकों का मूल्यांकन प्रस्तुत करेंगें । यह मूल्यांकन एक मानव द्वारा तथा कुछ स्वचालित मूल्यांकन मेट्रिक्स द्वारा किया जायेगा, जो वाक्य, दस्तावेज़ और प्रणाली स्तर पर होगा। अंत में हम इन दोनों मूल्यांकनो की तुलना भी करेंगें ।

*सूचक-शब्द:* मानव मूल्यांकन, स्वचालित मूल्यांकन, ब्लू मैट्रिक, मेटीयोर मैट्रिक

## 1. परिचय

जब से मशीन अनुवादक के विकास एवं अनुसन्धान की प्रक्रिया शुरू हुई है तभी से मशीनी अनुवाद का मूल्यांकन चल रहा है। शुरू में मूल्यांकन सिर्फ आदमीयों द्वार ही होता था। अब यह स्वचालित भी होता है। जब भी मशीनी अनुवादक में कोई नया विकास होता है तो एक मशीनी अनुवादक के प्रयोजना प्रबंधक के लिए यह जानना बहुत ज़रुरी होता है कि उसके अनुवादक में पिछले संस्करण से अभी तक कितना विकास हुआ। दुर्भाग्यवश यह प्रश्न इतना सीधा नहीं है क्योकि किसी मशीनी अनुवादक के दो संस्करणों में से दोनों ही बहुत अच्छा या बहुत बुरा अनुवाद दै सकते हैं या दोनों ही आंशिक रूप से सही अनुवाद दे सकते हैं जो अलग अलग तरीके से सही हो।

इसलिए मूल्यांकन बहुत जरुरी है जो स्पष्ट दिशा निर्देशों को क्रियान्वित करे। मानवीय मूल्यांकन की प्रक्रिया पिछले छह दशकों से चली आ रही है। इस प्रक्रिया में एक या अधिक मानवों द्वारा मूल्यांकन किया जाता है। मिल्लर और बीबी-सेंटर (1956) तथा पफ्फिन (1965) पहले ऐसे वज्ञानिक थे जिनने मानव द्वारा मशीनी अनुवादक के मूल्यांकन को सुझाया। तभी से, इस प्रकिया को सुदृड़ करने के लिए कई सारे अध्ययन किये गए है। इस लेख में हम ऐसी ही कुछ पद्धतियों (मानवीय तथा स्वचालित) का अध्ययन करेंगे। इस लेख के दूसरे खंड में हम पिछली कुछ पध्दितयो का अध्ययन करेंगे। तीसरे खंड में हम हमारी मूल्यांकन पद्धति का वर्णन करेंगे। चौथे खंड में हम इस पद्धति के द्वारा कुछ ऑनलाइन मशीनी अनुवादकों के मूल्यांकन के परिणामों की समीक्षा करेंगे तथा पांचवे खंड में हम इस प्रक्रिया का निष्कर्ष प्रस्तुत करेगें।

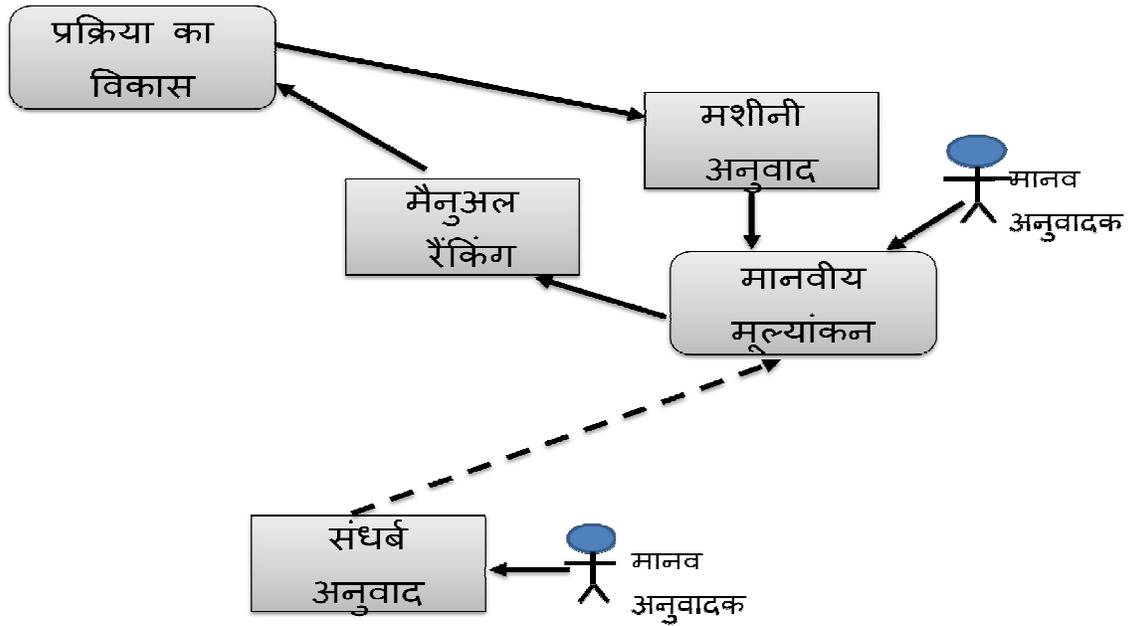

**चित्र 1:** मानवीय अनुवाद की प्रक्रिया

## 2. सम्बंधित कार्यों की समीक्षा

एलपेक (1956) ने पहला मशीनी अनुवाद किया था। उन्होंने इस कार्य के लिए कुछ इंसानों की मदद ली थी तथा उस समय के मशीनी अनुवादकों की गुणवत्ता का मूल्यांकन किया था। उन्होंने पाया की मशीनी अनुवाद की प्रक्रिया बहुत जटिल है तथा इसके द्वारा दिया गए अनुवाद अत्यंत खराब है तथा उन्होंने अमरीकी सरकार के रक्षा मंत्रालय, जो इस परियोजना के अनुसन्धान का निधिकरण कर रहा था, को सलाह दी की मशीनी अनुवादक में निधी देने की बजाय कुछ कम जटिल कम्प्यूटेशनल भाषाविज्ञान के कार्यों में पैसे का निवेश किया जाये। मशीनी अनुवादक के मुल्यांकन में एक बड़ी सफलता स्ल्यप (1979) में हासिल की गयी थी जिन्होंने सिसट्रांस नामक एक मशीनी अनुवादक के अनुवादकों की सुविकर्यता का अध्ययन किया। उन्होंने पाया कि मशीनी अनुवाद अपने आप एक इंसान जैसा अनुवाद नहीं कर सकता पर यदि एक इंसान को, जिसको अनुवाद का कार्य सोपा गया हो, यह अद्कचरे से अनुवाद दिए जाए तो वह कम समय में ज्यादा अनुवाद कर साकता है। इस अध्यन के खुलासे के बाद से एक मशीनी अनुवादक को एक संपादन उपकरण की तरह देखा जाने लगा। एस्क और होरी (2005) ने मशीनी अनुवाद के साथ उसके स्रोत वाक्य के अर्थ की तुलना का अध्ययन किया। उन्होंने पाया कि यही स्रोत वाक्य और अनुवादित

वाक्य, दोनों एक ही अर्थ को दर्शाते है तो हम अनुवाद को सही मान सकते हैं। गेट्स (1996) ने अर्थ सम्बंधी पर्याप्तता का अध्यन किया। उन्होंने दुभाषिय मानव अनुवादकों की सहायता से इस मूल्यांकन को किया। इस मूल्यांकन में उन्होंने अनुवादक को स्रोत एवं अनुवादित वाक्यो को पढ़ने को कहा तथा उनको बहु बिंदु स्तर पर रेट करने को कहा।

स्वचालित मूल्यांकन स्वतः अनुवाद की गुणवत्ता को निर्धारित करने का एक तरीका है। यह मूल्यांकन मानवीय मूल्यांकन से भिन्न होता है। इस मूल्यांकन में हमे मानव द्वारा किये गए अनुवादों की जरुरत होती है। हम इस अनुवाद की प्रक्रिया का मशीनी अनुवाद के साथ मिलान भी कर सकते हैं। चित्र 1 में मानवीय मूल्यांकन दर्शाया गया है तथा चित्र 2 में मशीनी अनुवाद तथा उसका मानवीय मूल्यांकन से मिलान दर्शाया गया है।

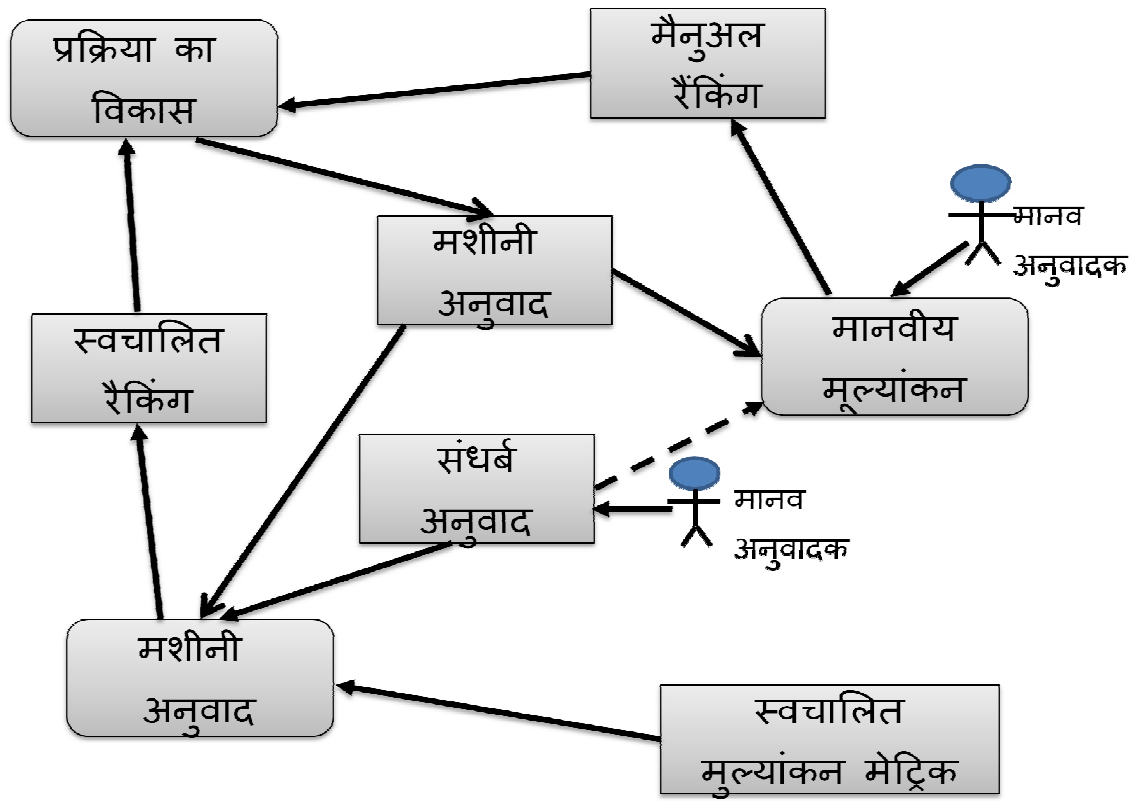

**चित्र 2**: मशीनी अनुवाद की प्रक्रिया

ब्लू (पपिनेनी 2001) पहली स्वचालित मैट्रिक थी जिसने मशीनी अनुवाद के मुल्यांकन में क्रांति लाने की कोशिश की। इसमें मानव द्वारा रचित संधर्भ अनुवाद को मशीनी अनुवाद से मिलाया जाता है जो 1...4 शब्दों का समूह होता है। इस मैट्रिक में ब्रेविटी पेनाल्टी नामक एक उपाय का उपयोग किया गया है जिसके द्वारा अगर मशीनी

अनुवाद मानवीय अनुवाद से छोटा हो तो उसे दण्डित किया जाता है इस मैट्रिक की गणना गाडना निम्नलिखित सूत्र द्वारा की जाती है।

$$\text{BLEU} = \min\left(1, \frac{\text{MT output-length}}{\text{reference-length}}\right) \times \exp\left(\sum_{n=1}^{N} w_n \log(p_n)\right) \quad (1)$$

निस्ट (डोगिंगटन 2002) इस मैट्रिक का संशोधित रूप है। इसका विकास राष्ट्रीय मानक एवं प्रोद्योगिकी संस्थान (निस्ट) द्वारा किया गया है, जिसके नाम पर इस मैट्रिक का नाम रखा गया। अपने एन-ग्राम के स्कोर की औसत की गणना ही निस्ट एवं ब्लू को भिन्न बनाती है। ब्लू ज्यामितीय मीन से अपना अंतिम स्कोर निकलती है, निस्ट सामानांतर मीन से अपना अंतिम स्कोर निकालती है।

मेटीयोर मैट्रिक (डेनकोवसकी एवं लेवी 2011) ब्लू की खामियों को दूर करने के लिए बनाई गयी है। इस मैट्रिक में, मशीनी अनुवाद एवं मानवीय संधर्भ अनुवाद में कई तरह के मिलान किये जाते है जो निम्न प्रकार है :-

1. **शब्दिक मिलान** - यहाँ ऐसे शब्दों का मिलान किया जाता है जो मशीनी अनुवाद तथा संधर्भ अनुवाद में समान हो।
2. **मूलशब्द मिलान** - यहाँ ऐसे मूल शब्दों का मिलान किया जाता है जो मशीनी अनुवाद तथा संधर्भ अनुवाद में समान हो।
3. **पर्यायवाची मिलान** - यहाँ पर्यायवाची शब्दों का मिलान किया जाता है जो मशीनी अनुवाद तथा संधर्भ अनुवाद में सामन हो।
4. **पैराफ़्रेज़ मिलान** - यहाँ ऐसे सांकेतिक शब्दों का मिलान किया जाता है जो मशीनी अनुवाद तथा संधर्भ अनुवाद में सामन हो।

यहाँ मिलान अलग-अलग स्तर पर होते है। हर स्तर पर उन शब्दों का मिलान होता है जिनका मिलान पिछले स्तर पर नहीं हुआ हो। क्रम बिंदु 1-3 में केवल एक शब्द का ही मिलान होता है जबकि क्रम बिंदु 4 में एक या उससे अधिक शब्दों का मिलान होता है।

## 3. मूल्यांकन परिक्रिया

हमनें एक हज़ार वाक्यों का कोश तैयार किया है। यह वाक्य पर्यटन डोमेन से लिए गए है जिनहें दस दस्तावेजों में व्यवस्थित किया गया है। हर दस्तावेज में सौ-सौ वाक्य है। इस कोश का हमने तीन मशीनी अनुवादकों पर परिक्षण किया है। ये मशीनी अनुवादक है:

1. **गूगल अनुवादक** - यह अनुवादक सबसे लोकप्रिय अनुवाद है जो मुफ्त में अनुवाद प्रदान करता है । इस अनुवादक को गूगल कार्पोरेशन ने बनाया है।
2. **बिंग अनुवादक** - यह अनुवादक तेजी से गूगल की जगह ले रहा है। इस अनुवादक को मिक्रोसोफ्ट कार्पोरेशन ने बनाया है।

3. **ईबीएमटी (जोशी व अन्य 2010)** - यह अनुवादक हमने बनाया है। इस अनुवादक को हमने अपनी मशीनी अनुवादकों की तकनीकी समझ को विकसित करने के लिए बनाया है।

मूल्यांकन के लिए हमने अंग्रेजी-हिंदी भाषा युग्म का प्रयोग किया है। हमनें मानवीय तथा स्वचालित मूल्यांकन का प्रयोग किया। स्वचालित मूल्यांकन के लिए हमने ब्लू व मेटियोर मैट्रिक का प्रयोग किया तथा हमने इस मूल्यांकन को वाक्य स्तर पर केन्द्रित किया। हमने इन दोनों मै मैट्रिकों का परिणाम एक तथा चार संधर्भ वाक्यों के साथ पंजीकृत किया।

क्योंकि ब्लू पहली स्वचालित मैट्रिक थी तथा किसी भी मूल्यांकन की इस मैट्रिक के बिना कल्पना नहीं की जा सकती। हम हिंदी भाषा पर ब्लू के असर को देखना चाहते थे इसलिए भी हमने ब्लू को प्रयोग में लिया। मेटियोर का प्रयोग किया गया क्योंकि हम सतही भाषाई रूपों का हिंदी पर प्रभाव देखना चाहते थे। यहाँ मूल-शब्द मिलान के लिए हमने एक लाइटवेइट स्टेमर का प्रयोग किया जो रंगनाथन व राव (2003) द्वारा रचित एल्गोरिथ्म पर आधारित है तथा पर्यायवाची मिलान के लिए हमने वर्डनेट (नारायण व अन्य 2008) का प्रयोग किया। एक बहुत बुनियादी मूल्यांकन (जोशी व अन्य 2012) में यह पाया गया था की ब्लू हिंदी पर बहुत अच्छा परिणाम नहीं देती है। इसलिए वर्तमान मूल्यांकन में हमने ब्लू तथा मेटियोर दोनों के ही कई संस्करणों के साथ प्रयोग किया।

मानवीय मूल्यांकन के लिए हमने एक मैट्रिक का अविष्कार किया जो दस बिन्दुओ पर मूल्यांकन करती है। यह दस बिंदु है :-

1. संज्ञाओं के लिंग व वचन का अनुवाद में प्रयोग।
2. मूल वाक्य में प्रयुक्त काल का अनुवाद में प्रयोग।
3. मूल वाक्य में प्रयुक्त वाच्य का अनुवाद में प्रयोग।
4. व्यक्तिवाचक संज्ञा की पहचान।
5. विशेषण व क्रिया विशेषण का मूल वाक्य में संज्ञा व क्रिया के अनुकूल प्रयोग।
6. अनुवाद में सही शब्दों/पर्याय का चयन।
7. अनुवाद में संज्ञा, क्रिया एवं सहायक क्रिया का क्रम।
8. अनुवाद में विराम चिन्हों का प्रयोग।
9. अनुवाद में मूल वाक्य में प्रयुक्त महत्वपूर्ण भाग पर बल।
10. अनूदित वाक्य में मूल वाक्य में निहित अर्थ का सही समागम।

मनुष्यों को एक स्रोत वाक्य तथा उसका मशीनी अनुवाद दिया जाता है और उनसे इस दोनों को पड़ने के बाद इन दस बिंदुओं की 0-4 में रेटिंग करने को कहा जाता है। अंत में इन सभी रेटिंगों का औसत निकाल कर अंतिम स्कोर प्राप्त किया जाता है।

## 4. मुल्यांकन का परिणाम

हमने हर मशीनी अनुवादक को ब्लू व मेटियोर के चार-चार संस्करणों पर मूल्यांकित किया। मानवीय मूल्यांकन में गूगल और बिंग को सर्वश्रेष्ठ पाया गया। हमारा प्रयास इस मूल्यांकन को स्वचालित मूल्यांकन के ज़रिये दर्शाना है। इनके लिए हमने परिणामों का मानवीय मूल्यांकन के साथ स्वचालित मूल्यांकन का मिलान भी किया। इस प्रयोग के परिणाम टेबिल 1 में विदित है।

| | गूगल | | बिंग | | ईबीएमटी | |
|---|---|---|---|---|---|---|
| | एक संधर्ब वाक्य | चार संधर्ब वाक्य | एक संधर्ब वाक्य | चार संधर्ब वाक्य | एक संधर्ब वाक्य | चार संधर्ब वाक्य |
| **ब्लू 1-ग्राम** | 0.050 | 0.099 | 0.073 | 0.108 | **0.094** | **0.110** |
| **ब्लू 2-ग्राम** | 0.062 | 0.099 | 0.073 | 0.111 | **0.115** | **0.141** |
| **ब्लू 3-ग्राम** | 0.074 | 0.113 | 0.068 | 0.089 | **0.105** | **0.125** |
| **ब्लू 4-ग्राम** | 0.084 | **0.118** | 0.077 | 0.106 | **0.106** | 0.117 |
| **मेटियोर शाब्दिक व मूल-शब्द मिलान** | **0.108** | 0.087 | 0.065 | 0.098 | 0.100 | **0.107** |
| **मेटियोर शाब्दिक व पर्यायवाची मिलान** | 0.007 | 0.064 | 0.065 | **0.120** | 0.109 | 0.114 |
| **मेटियोर शाब्दिक, मूल-शब्द व पर्यायवाची मिलान** | 0.014 | 0.053 | 0.063 | **0.118** | **0.111** | 0.096 |
| **मेटियोर शाब्दिक, मूल-शब्द, पर्यायवाची व पैराफ़्रेज़ मिलान** | 0.019 | 0.011 | 0.007 | **0.088** | **0.040** | 0.048 |

**टेबिल 1**: मशीनी अनुवाद का स्वचालित मूल्यांकन के साथ मिलान (correlation)

इस अध्यन में, ईबीएमटी ने ब्लू के सभी संस्करणों के साथ, एक तथा चार संधर्भ वाक्यों में अपनी दक्षता साबित की। इसमें केवल एक अपवाद रहा जिसमें ब्लू ४-ग्राम, चार वाक्यों के साथ, गूगल ने सबसे अच्छा प्रदर्शन किया। मेटियोर के परिणाम ब्लू से थोड़े भिन्न थे। इसमें मेटियोर शाब्दिक व मूल-शब्द मिलान में एक वाक्य के साथ गूगल ने अच्छा प्रदर्शन किया और चार वाक्य के साथ ईबीएमटी ने अच्छा प्रदर्शन किया। मेटियोर शाब्दिक व पर्यायवाची मिलान, एक वाक्य के साथ ईबीएमटी ने अच्छा प्रदर्शन किया तथा चार वाक्यों के साथ बिंग ने अच्छा

प्रदर्शन किया। ये प्रचलन मेटियोर शाब्दिक, मूल-शब्द व पर्यायवाची मिलान और मेटियोर शाब्दिक, मूल-शब्द, पर्यायवाची व पैराफ़्रेज़ मिलान के साथ, दोनों, एक संधर्भ वाक्य और चार संधर्भ वाक्यों में भी देखा गया।

## 5. निष्कर्ष

इस लेख में हमने तीन अंग्रेज़ी-हिंदी मशीनी अनुवादकों के मूल्यांकन का विश्लेषण किया है। इस विश्लेषण में हमने मानवीय मूल्यांकन का स्वचालित मुल्यांकन के साथ मिलान भी किया है। हमने ये पाया की ब्लू कई बार सही मूल्यांकन नहीं कर पाती। ऐसा इसलिए भी हो सकता है क्योकि ब्लू का सैधांतिक आधार है की किसी भी अच्छे स्रोत वाक्य के अनुवाद भी अच्छे होगे। यह शायद प्राकृतिक भाषाओं के अर्थपूर्णता और निहित अस्पष्टता के कारण नहीं हो सकता है।

इसके बनिस्पत, मेटियोर ने अच्छे परिणाम दिए, लेकिन अक्सर, जब सिर्फ एक ही संधर्भ वाक्य के साथ स्वचालित मूल्यांकन किया गया तब ये मैट्रिक ठीक परिणाम नहीं दे पाई। इसका एक कारण, इस मैट्रिक का सतही भाषा वैज्ञानिक स्तर होना है। अगर हम और अधिक जयादा भाषा वैज्ञानिक स्तर पर थोड़ा और गहन अध्ययन करें तो शायद हम मानवीय मूल्याँकन जैसे ही परिणाम मिले। इस समय हम सिर्फ यही निष्कर्ष निकाल सकते है कि मेटियोर शाब्दिक व पर्यायवाची मिलान या मेटियोर शाब्दिक, मूल-शब्द व पर्यायवाची मिलान या मेटियोर शाब्दिक, मूल-शब्द, पर्यायवाची व पैराफ़्रेज़ मिलान का मुल्यांकन अगर चार वाक्यों के साथ किया जाए तो मानवीय मूल्यांकन के जैसे ही परिणाम मिल सकते है।

## संधर्भ